\documentclass[letterpaper]{article} 
\usepackage[submission]{aaai2026}  
\usepackage{times}  
\usepackage{helvet}  
\usepackage{courier}  
\usepackage[hyphens]{url}  
\usepackage{graphicx} 
\usepackage{natbib}  
\usepackage{caption} 
\usepackage{algorithm}
\usepackage{algorithmic}
\usepackage{booktabs}   
\usepackage{caption}    
\usepackage{amssymb}    
\usepackage{multirow}   
\usepackage{makecell}   
\usepackage{xcolor} 
\usepackage{float}

\usepackage{graphicx} 
\usepackage{amssymb} 

\usepackage{newfloat}
\usepackage{listings}
\DeclareCaptionStyle{ruled}{labelfont=normalfont,labelsep=colon,strut=off} 
\floatstyle{ruled}
\newfloat{listing}{tb}{lst}{}
\floatname{listing}{Listing}
\title{A Federated Learning Framework for Handling Subtype Confounding and Heterogeneity in Large-Scale Neuroimaging Diagnosis}

\affiliations{
    \textsuperscript{\rm 1}Association for the Advancement of Artificial Intelligence\\

    1101 Pennsylvania Ave, NW Suite 300\\
    Washington, DC 20004 USA\\
    proceedings-questions@aaai.org
}

\usepackage{bibentry}
\makeatletter

\def\@maketitle{%
  \newpage \null \vskip 2em%
  \begin{center}%
  {\LARGE\bf \@title \par}%
  \vskip 1.5em%
  {\large                
    \begin{tabular}[t]{c}%
      Xinglin Zhao\textsuperscript{1},%
      \;Yanwen Wang\textsuperscript{1},%
      \;Xiaobo Liu\textsuperscript{2},%
      \;Yanrong Hao\textsuperscript{1},%
      \;Rui Cao\textsuperscript{1},%
      \;Xin Wen\textsuperscript{1,*}%
    \end{tabular}\par}%
  \vskip 1em%
  {\small
    \textsuperscript{1}Taiyuan University of Technology \\
    \textsuperscript{2}Montreal Neurological Institute Mcgill University \\
    \texttt{Xinglin Zhao: 2024511307@link.tyut.edu.cn \\ Xin Wen: xwen@tyut.edu.cn}}
  \end{center}%
  \vskip 1.5em}
\makeatother

\begin{document}

\maketitle

\begin{abstract}
Computer-aided diagnosis (CAD) systems play a crucial role in analyzing neuroimaging data for neurological and psychiatric disorders. However, small-sample studies suffer from low reproducibility, while large-scale datasets introduce confounding heterogeneity due to multiple disease subtypes being labeled under a single category. To address these challenges, we propose a novel federated learning framework tailored for neuroimaging CAD systems. Our approach includes a dynamic navigation module that routes samples to the most suitable local models based on latent subtype representations, and a meta-integration module that combines predictions from heterogeneous local models into a unified diagnostic output. We evaluated our framework using a comprehensive dataset comprising fMRI data from over 1300 MDD patients and 1100 healthy controls across multiple study cohorts. Experimental results demonstrate significant improvements in diagnostic accuracy and robustness compared to traditional methods. Specifically, our framework achieved an average accuracy of 74.06\% across all tested sites, showcasing its effectiveness in handling subtype heterogeneity and enhancing model generalizability. Ablation studies further confirmed the importance of both the dynamic navigation and meta-integration modules in improving performance. By addressing data heterogeneity and subtype confounding, our framework advances reliable and reproducible neuroimaging CAD systems, offering significant potential for personalized medicine and clinical decision-making in neurology and psychiatry.
\end{abstract}


\section{Introduction}
Computer-aided diagnosis (CAD) has become an indispensable tool in the analysis of functional magnetic resonance imaging (fMRI) data, enabling the identification of complex patterns of brain activity associated with neurological and psychiatric disorders\cite{fcST,dlinns,multisite}. As the field transitions toward more data-driven and personalized approaches, CAD systems are playing a pivotal role in improving early detection, differential diagnosis, and treatment monitoring\cite{mdd,multisite}.

However, a major limitation in current neuroimaging research is the reliance on small-sample studies, which often suffer from low statistical power and poor reproducibility\cite{datasets}. It has been widely recognized that robust and reproducible brain-wide association studies require datasets encompassing thousands of individuals\cite{moresamples}. While increasing the sample size enhances statistical reliability, it also introduces new challenges. Notably, aggregating data across multiple institutions or populations often results in increased heterogeneity — for instance, a single diagnostic label may encompass multiple subtypes of a disorder\cite{bigdata}. This confounding variability can mislead conventional CAD models, leading to reduced generalizability and compromised diagnostic performance.

To address these challenges, federated learning (FL) has emerged as a promising paradigm for distributed neuroimaging analysis\cite{fedbrainimage,fedhealth,fedsurvey}. In this setup, each participating site trains a local model on its own subset of data , potentially representing a specific disease subtype, while a central server coordinates the learning process to ensure global consistency. This approach enables collaborative learning without centralizing sensitive patient data, thereby preserving privacy and complying with regulatory requirements. However, applying FL to neuroimaging CAD systems presents two key challenges. First, determining the appropriate local model for a given sample, especially when the global label space is shared across multiple subtypes, remains an open problem, referred to as the navigation problem. Second, integrating heterogeneous local models into a unified and robust diagnostic system without compromising performance is non-trivial.

In this work, we propose a novel federated learning framework specifically designed for neuroimaging CAD systems that effectively addresses both the navigation and integration challenges in the presence of subtype heterogeneity. Our approach introduces a dynamic routing mechanism to assign samples to the most suitable local models during training, and a meta-integration strategy to combine predictions across models into a coherent and reliable diagnostic output. By preserving data privacy and enhancing model generalizability, our framework enables more accurate and reproducible computer-aided diagnosis in large-scale, heterogeneous neuroimaging studies. The innovations of this paper can be summarized as follows:
\begin{itemize}

\item \textbf{Dynamic Navigation Mechanism:} A dynamic routing mechanism is proposed to assign samples to the most suitable local models for training based on latent subtype representations. This mechanism addresses the challenge of selecting the appropriate model when dealing with multiple disease subtypes, thereby enhancing diagnostic accuracy.

\item \textbf{Meta-Integration Strategy:} An innovative meta-integration approach has been introduced to effectively combine predictions from heterogeneous local models into a unified diagnostic output. This method not only tackles the challenges posed by data heterogeneity but also ensures robustness and consistency of the diagnostic system.

\item \textbf{Two-Stage Framework Based on Adaptive Attention Aggregation (AAA):} A two-stage framework tailored for brain image classification has been developed. The first stage involves site feature learning and heterogeneous classification to capture differences in multi-site medical data. The second stage utilizes a cross-site model adaptive aggregation based on an attention mechanism to achieve personalized prediction at the individual sample level. This framework significantly improves the system's performance in handling data heterogeneity and subtype confounding, providing more accurate diagnosis.

\end{itemize}

\section{Related Work}
\subsection{Heterogeneous in fMRI-based CAD}
Among numerous methods for fMRI-based CAD in neurological and psychiatric disorders, heterogeneous are caused by confounding factors including device diversity, individual differences, and potential subtypes, and balancing them is a key challenge\cite{confounderfree}. Many works have addressed heterogeneity by adversarially separating the primary classification task from site characteristics and individual variability. For example, attention mechanisms are constructed using non-imaging data representations to suppress individual differences\cite{nonimage}; statistical methods (Subsampling Maximum-mean-distance and CovBat) are employed to correct site-specific variations\cite{covbat}; and domain adaptation approaches map data into a target domain to alleviate both individual and site discrepancies\cite{domdd,dotpami}. Such works typically treat site information and individual differences as confounders, but do not address potential disease subtypes. Moreover, they rely on manually designed features for confounder representation, lacking an automatic representation learning scheme. Additionally, domain adaptation methods require pre-specifying a target domain.

Moreover, MDD encompasses multiple subtypes with low discriminative power between them, which can introduce significant confusion into CAD systems. Effectively addressing disease subtypes and developing an automatic confounder representation learning method to improve CAD performance remains an unexplored area.
\subsection{Federated Learning}
In real-world applications, data are often distributed across decentralized sites with distinct feature distributions due to device variations, patient populations, or acquisition protocols which lead to significant data heterogeneity~\cite{hetero,fedgh}. This non-IID nature poses challenges for centralized training, motivating the use of FL, where clients train locally and exchange only model updates, preserving data privacy while enabling collaboration~\cite{2016, FL, mode}.

In medical imaging, FL has emerged as a promising approach for building CAD systems without violating privacy regulations~\cite{muti,pfedes}. To handle inter-site heterogeneity, personalized FL (PFL) methods have been proposed~\cite{pfedafm, fedssa}, which learn site-specific models or deploy local classifiers~\cite{fedwm, ffedcl}. These approaches improve performance by adapting global knowledge to local data characteristics, showing strong potential in clinical settings.

However, most existing PFL methods perform personalization at the client level, assuming uniformity within each site. This ignores intra-site variability and limits adaptability at inference time. In contrast, this work proposes AAA, a framework that enables sample-level personalization through a dynamic navigation module and heterogeneous classifiers, allowing fine-grained adaptation to diverse inputs and better handling of complex data and model heterogeneity in medical FL.

\section{Methodology}
\subsection{Overview}
We propose Adaptive Attention Aggregation (AAA), a two-stage framework for brain image classification that addresses data heterogeneity in multi-center medical data and explores the impact of disease subtypes on diagnostic accuracy. AAA enables personalized, sample-level predictions through heterogeneous modeling and adaptive fusion, as illustrated in Figure 1.

\textbf{Stage I: Site-Specific Feature Learning and Heterogeneous Classification.}

At each participating site, we deploy a shared homogeneous autoencoder and a site specific heterogeneous classifier. The autoencoder learns a common low-dimensional representation of local neuroimaging data, promoting cross-site compatibility. Concurrently, it generates two class-specific prototype templates that capture the intrinsic data structure of the site. The heterogeneous classifier, tailored to each site’s data distribution, performs local classification using these representations. After local training, each site uploads its autoencoder parameters, classifier weights, and learned prototype templates to the central server.

\textbf{Stage II: Attention-Based Adaptive Aggregation.}

During inference, a test sample is first encoded into its low-dimensional representation using the shared autoencoder. The system then computes the cosine similarity between this representation and each site’s prototype templates, producing attention weights that reflect the compatibility between the input and each site’s data distribution. These weights are used within a Mixture-of-Experts (MoE) mechanism to dynamically aggregate the predicted logits from all site-specific classifiers. This adaptive fusion enables sample-level personalization, where the final prediction is informed by the most relevant local models based on distributional similarity.

By integrating subtype-aware modeling and attention-driven aggregation, AAA effectively mitigates the impact of data heterogeneity in multi-center settings. The results demonstrate that modeling subtypes improves classification accuracy, and the proposed framework significantly enhances model performance through personalized, context-aware inference.
\begin{figure*}[htb]
    \centering
    \includegraphics[width=0.8\textwidth]{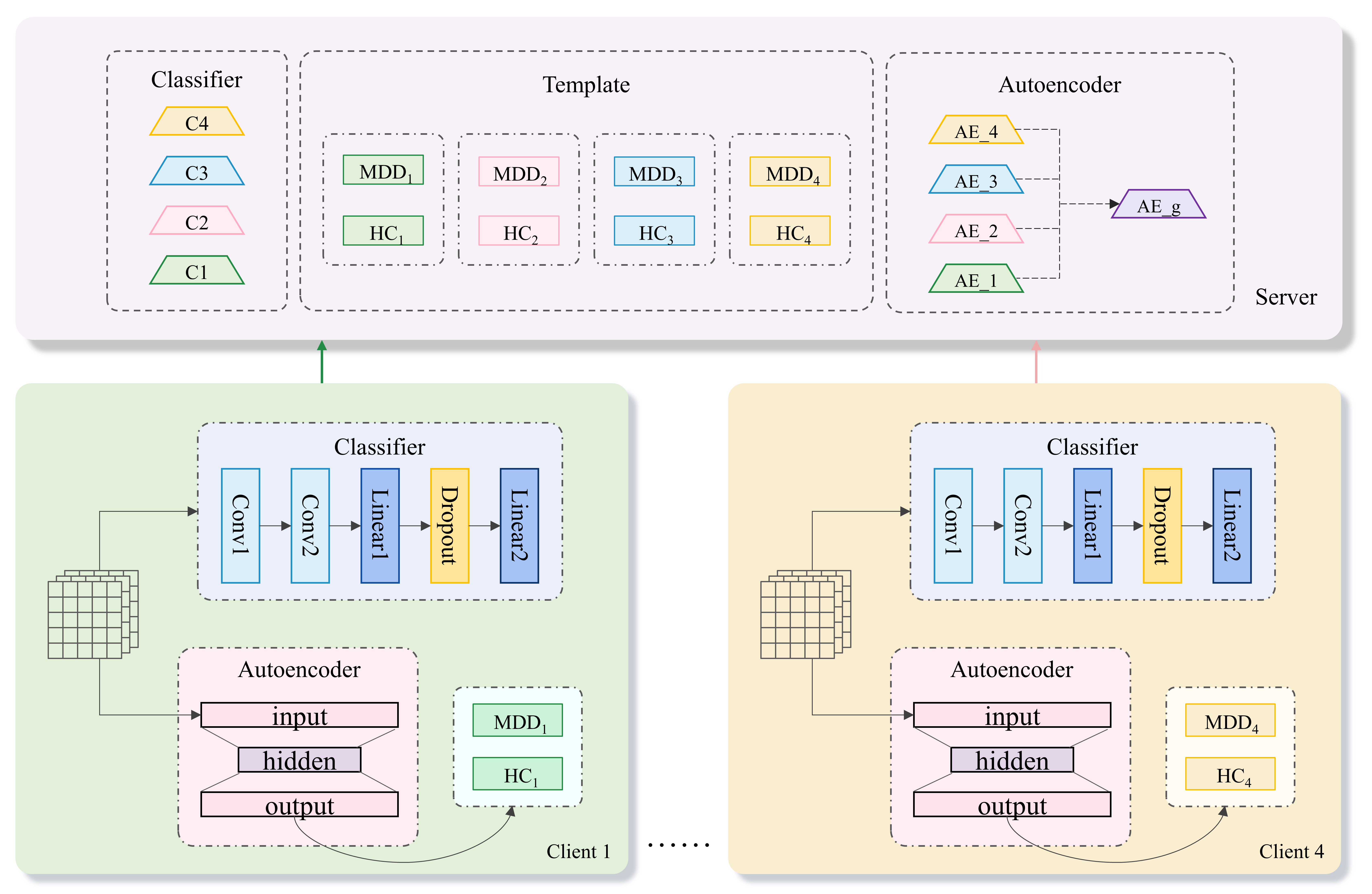} 
    \caption{Overall framework of the proposed AAA method}
    \label{fig:1}
\end{figure*}

\subsection{Notations and Definitions}

We first clarify the notations used for analysis in Table 1.


\begin{table}[htb]
  \centering
  \caption{Description of Notations}
  \label{Notations}
  \begin{tabular}{@{}>{\raggedright\arraybackslash}p{3cm}p{5cm}@{}}
    \toprule
    \textbf{Notation} & \textbf{Description} \\
    \midrule
    $N$ & Total number of sites \\
    $s \in 1, 2,\dots, N$ & Local site \\
    $E_{s}$ & Autoencoder for site $s$ \\
    $C_{s}$ & Classifier for site $s$ \\
    $y\in \{0,1\}$ & Data category of site \\
    $\mathbf{S}$ & Vector reconstructed by autoencoder \\
    $\mathbf{T}$ & Low dimensional representation extracted by autoencoder \\
    $\bar{\mathbf{T}}_{s,y}$ & Latest template for the sample \\ 
    & with label $y$ at site $s$ \\           
    \bottomrule
  \end{tabular}
\end{table}


\subsection{Stage I: Site feature learning and heterogeneous classification}
To effectively address the inherent data heterogeneity issue in multi-site medical data, enhance the local classification performance of each site, and gain a deeper understanding of the data differences between sites, we have designed a site feature learning and heterogeneous classification module. This module receives labeled datasets $\{\mathcal{D}_1, \mathcal{D}_2, ..., \mathcal{D}_N\}$ from $N$ different sites as input. The core architecture consists of a set of homogeneous autoencoders $\{E_{1}, E_{2}, ..., E_{N}\}$ and a set of heterogeneous classifiers $\{C_{1}, C_{2}, ..., C_{N}\}$. Through the homogeneous autoencoder, we obtain the template for each site, including the MDD template and the NC template. Finally, the heterogeneous classifier $C_{s}$, homogeneous autoencoder $E_{s}$, and template $T_{s}$ for each site are uploaded to the server. The server aggregates the homogeneous autoencoders $E_{s}$ from each site with weights to obtain the global homogeneous autoencoder $E_{g}$, while storing the heterogeneous classifiers $C_{s}$ and template $T_{s}$ for each site.
The workflow of Stage I is shown in Figure~\ref{fig:relative-path}.

\begin{figure}[htb]
    \centering
    \includegraphics[scale=0.4]{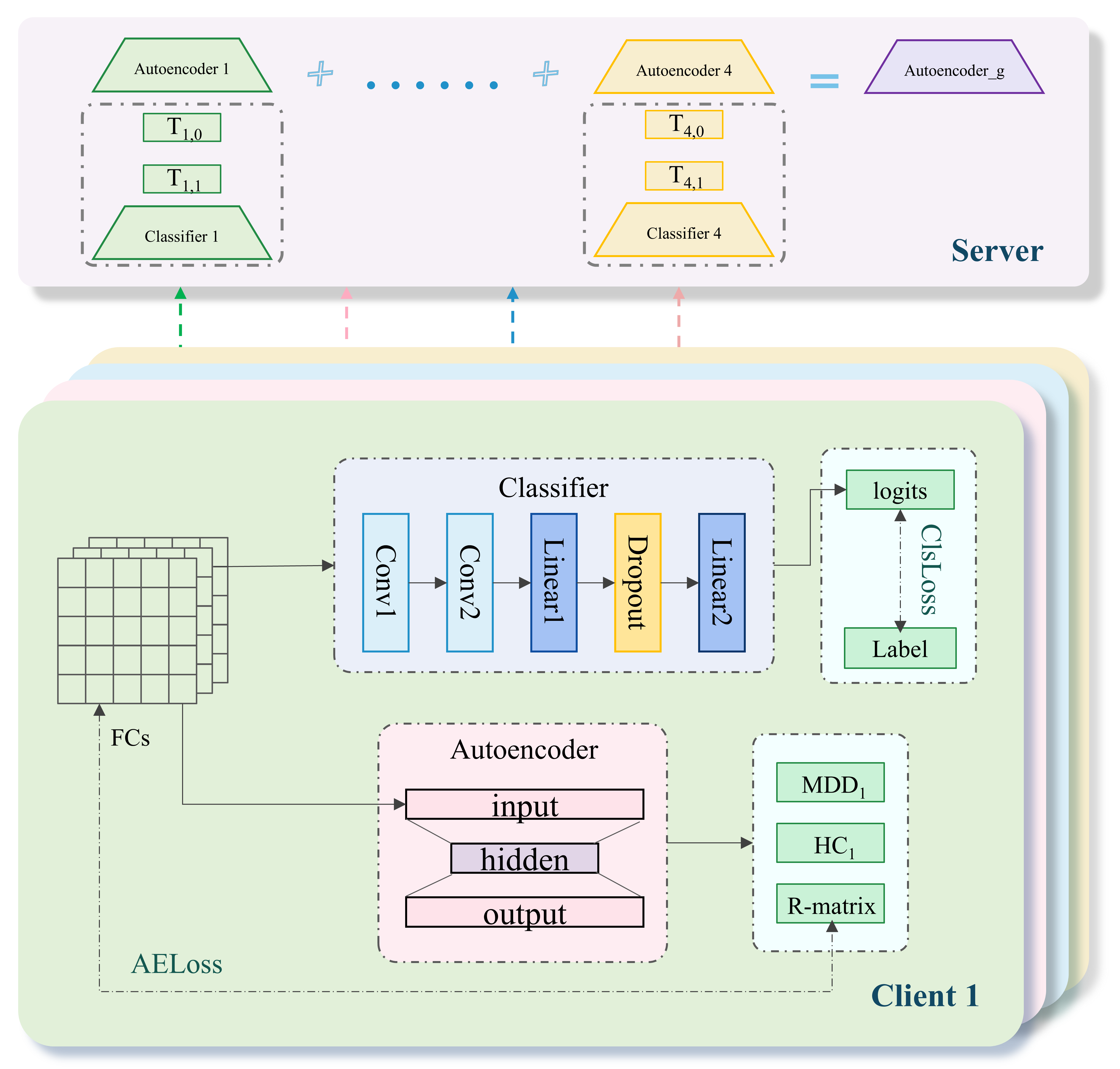} 
    \caption{Workflow of Stage I}
    \label{fig:relative-path}
\end{figure}

To capture the overall distribution of FC corresponding to different diagnostic labels within each site, we designed an autoencoder, denoted as:
\begin{equation}\mathbf{S,T}=\mathcal{E}_\theta(\mathbf{x}),\mathbf{x}\in\mathbf{R}^d, d=\frac{n(n-1)}{2},\end{equation}
where $\mathcal{E}_\theta$ denotes the encoding-decoding network with parameter $\theta$, $\mathbf{x}$ denotes the upper triangular flattened vector of a single sample (with length $d$, n is the number of ROIs), $\mathbf{S}$ denotes the reconstructed vector, $\mathbf{T}$ denotes the low dimensional representation. By utilizing the symmetry of the FC matrix to extract the upper triangular elements and construct feature vectors as input, the encoder part of the autoencoder maps the feature vectors to the latent feature space, obtains the low dimensional representation of the corresponding label, and averages it, further obtain the template for the label,and the mapping process can be formalized as follows:
\begin{equation}\bar{\mathbf{T}}_{s,y}=\frac{1}{|\mathcal{D}_{s,y}|}\sum_{\mathbf{x}_{i}\in\mathcal{D}_{s,y}}T_i,T_i=\mathcal{E}_{\theta}(\mathbf{x}_{i}), y\in\{0,1\},\end{equation}
where $\mathcal{D}_{s,y}$ denotes all sample sets with label $y$ at site $s$, $|\mathcal{D}_{s,y}|$ denotes the corresponding sample quantity, $\bar{\mathbf{T}}_{s,y}$ denotes the latest template for the sample with label $y$ at site $s$. The output of the autoencoder is the reconstructed upper triangular vector, and the autoencoder is continuously optimized by reducing the difference between the reconstructed data and the original data, and the optimization process can be formalized as follows:
\begin{equation}\mathcal{L}_{\cos}=1-\frac{\mathbf{S}\cdot\mathbf{x}}{\|\mathbf{S}\|_2\|\mathbf{x}\|_2}\end{equation}
where $S$ denotes the upper triangular vector output by the decoder,$\mathbf{S}\cdot\mathbf{x}$ denotes dot product, $|| . ||_2^2$ denotes $L_2$ norm. 
Record the latest $\bar{\mathbf{T}}_{s,0}$ and $\bar{\mathbf{T}}_{s,1}$ obtained, and upload the trained $E_s$, $\bar{\mathbf{T}}_{s,0}$, and $\bar{\mathbf{T}}_{s,1}$ to the server.

At the same time, a heterogeneous CNN classifier was designed to achieve end-to-end training of ROI functional connectivity matrices with a size of $n * n$. The CNN network fully considers the "row column two dimensional" semantics of the FC matrix, and CNN network can be formalized as follows:
\begin{equation}\hat{\mathbf{y}}=\mathcal{F}_\Theta(\mathbf{X}),\mathbf{X}\in\mathbf{R}^{n\times n},\hat{\mathbf{y}}\in\mathbf{R}^2,\end{equation}
where $\mathcal{F}_\Theta$ denotes the entire CNN network, parameter set $\Theta = \{conv,linear\}$, $\mathbf{X}$ denotes the input function connection matrix, $\hat{\mathbf{y}}$ denotes 2D softmax prediction vector. The input data is function connection matrix,we first expand the FC matrix to 3 dimensions, $i.e.$, $1 * n * n$, then pass through the first convolution layer with a kernel size of $1*n$ to extract row information, While preserving row correlations, the n dimensional features of each row are compressed into 1 dimension to obtain a row digest matrix, and the convolution process can be formalized as follows:
\begin{equation}\mathbf{Z}_1=\phi\left(\mathrm{IN}\left(\mathbf{W}_1*\mathbf{X}_{3\mathrm{D}}\right)\right),\mathbf{X}_{3\mathrm{D}}\in\mathbf{R}^{1\times n\times n},\mathbf{W}_1\in\mathbf{R}^{1\times n},\end{equation}
where $\mathbf{X}_{3\mathrm{D}}$ denotes 3D tensor after channel dimension expansion, $\mathbf{W}_1$ denotes the first layer of 2D convolution kernel has an output channel number of $c_1$ and a kernel size of $(1, n)$, $\mathrm{IN}( . )$ denotes $InstanceNorm2d$, $\phi$ denotes nonlinear activation function. After passing through the second convolution layer with a kernel size of $n * 1$, column information is extracted, and the entire FC matrix is finally compressed into a scalar of $1 * 1$, achieving complete compression in the spatial dimension,and the second convolution process can be formalized as follows:
\begin{equation}\mathbf{Z}_2=\phi\left(\mathbf{W}_2*\mathbf{Z}_1\right),\mathbf{W}_2\in\mathbf{R}^{n\times 1},\end{equation}
where $\mathbf{W}_2$ denotes the second layer of 2D convolution kernel has an output channel number of $c_2$ and a kernel size of $(n, 1)$, $\mathbf{Z}_1$ denotes the row digest matrix output from the first layer. The compressed $1 × 1$ scalar is gradually mapped to a low dimensional latent space through three fully connected layers, and the last layer outputs binary logits corresponding to the predicted probabilities of MDD or NC. Mapping process can be formalized as follows:
\begin{equation}{\mathbf{l}}=\mathrm{softmax}\left(\mathbf{W}_{\mathrm{out}}\cdot\mathrm{Drop}\left(\phi(\mathbf{W}_{\mathrm{hid}}\cdot\mathbf{Z}_2)\right)\right),\end{equation}  
where $\mathbf{W}_{\mathrm{hid}}$ denotes the hidden layer weight, $\mathbf{W}_{\mathrm{out}}$ denotes the output layer weight, $\mathrm{Drop}$ denotes the probability of random dropout. During the training process, we use cross entropy loss to measure the difference between predicted labels and real labels, and continuously optimize the classifier by reducing the difference, and the optimization process can be formalized as follows:
\begin{equation}\mathcal{L}=-\frac{1}{|\mathcal{D}_s|}\sum_{i=1}^{|\mathcal{D}_s|}\log\frac{\exp(z_{i,y_{i}})}{\exp(z_{i,0})+\exp(z_{i,1})},\end{equation}
where $y_{i}$ denotes the real label, $z_{i,0}$ denotes the sample $i$ corresponds to logits with label 0, $z_{i,1}$ denotes the sample $i$ corresponds to logits with label 1, and then upload the trained classifier to the server for storage.

Finally, calculate the weights of each site on the server. Calculation process can be formalized as follows:
\begin{equation}w_s=\frac{|\mathcal{D}_s|}{\sum_{k=1}^N|\mathcal{D}_k|},\end{equation}
where $w_s$ denotes Weight of site s, $N$ denotes the total number of sites. Weighted average the homogeneous autoencoders $E_i$ of each site to obtain the global homogeneous autoencoder $E_g$, and calculation process can be formalized as follows:
\begin{equation}\theta^\mathrm{global}=\sum_{s=1}^Nw_s\theta^{(s)},\end{equation}
where $\theta^{(s)}$ denotes local parameters of site s autoencoder, $\theta^\mathrm{global}$ denotes parameters of autoencoder after global aggregation.

In this module, each site trains local heterogeneous classifiers and homogeneous autoencoders with local data to obtain MDD template and NC template, and uploads heterogeneous classifier $C_i$, homogeneous autoencoder $E_i$, MDD template $\bar{\mathbf{T}}_{s,0}$, and NC template $\bar{\mathbf{T}}_{s,1}$ to the server. The server weight aggregates the homogeneous autoencoders of each site to obtain global autoencoder $E_g$, and distributes the global autoencoder $E_g$, heterogeneous classifier $C_i$, MDD template $\bar{\mathbf{T}}_{s,0}$ and NC template $\bar{\mathbf{T}}_{s,1}$ to each site, effectively addressing the heterogeneity challenges in multi center medical data analysis, improving the classification accuracy of the model, and enhancing the understanding of data distribution differences.

\subsection{Stage II: Cross-site model adaptive aggregation based on attention mechanism}
Firstly, the autoencoder $E_i$ trained in Stage I is used to process the upper triangular flattening vector of a single sample, obtaining the low dimensional representation $T$ of the sample. Then, the FC matrix is input to heterogeneous classifiers at each site, and the predicted logits $\mathbf{l}$ on each site model are obtained separately, calculate the attention score of each site, weight the logits of each site based on the attention score, and obtain the final logits value to achieve personalized prediction of a single sample.
The workflow of Stage II is shown in Figure 3.

\begin{figure}[htb]
    \centering
    \includegraphics[scale=0.4]{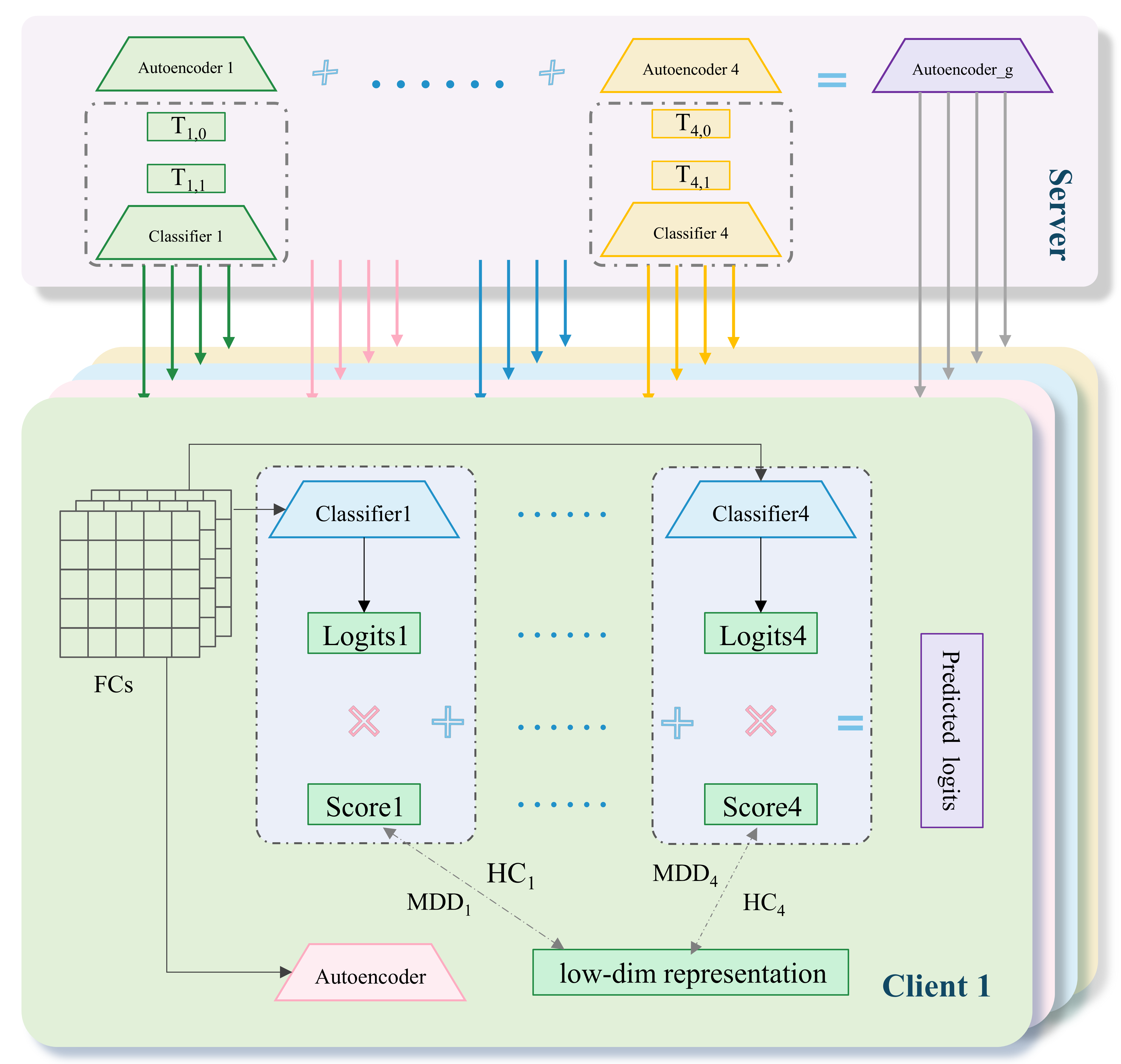} 
    \caption{Workflow of Stage II}
    \label{fig:relative-path}
\end{figure}

Calculate the cosine similarity between the low dimensional representation $T$ of the sample and the two templates ($\bar{\mathbf{T}}_{s,0}$ and $\bar{\mathbf{T}}_{s,1}$) at site $s$. Calculation process can be formalized as follows:

\begin{equation}\cos(T,\bar{\mathbf{T}}_{s,y})=\frac{T^\top\bar{\mathbf{T}}_{s,y}}{\|T\|\|\bar{\mathbf{T}}_{s,y}\|}=\frac{\sum_{i=1}^dT_i\bar{\mathbf{T}}_{s,y,i}}{\sqrt{\sum_{i=1}^dT_i^2}\sqrt{\sum_{i=1}^d\bar{\mathbf{T}}_{s,y,i}^2}},\end{equation}
where $d$ denotes the number of elements in the upper triangle, $\sum_{i=1}^dT_i\bar{\mathbf{T}}_{s,y,i}$ denotes the dot product of low dimensional representation $T$ and template $\bar{\mathbf{T}}_{s,y}$, used to measure the degree of alignment between the two in the direction, $\sqrt{\sum_{i=1}^dT_i^2}$ denotes the euclidean norm of low dimensional representation $T$, $\sqrt{\sum_{i=1}^d\bar{\mathbf{T}}_{s,y,i}^2}$ denotes the euclidean norm of template $\bar{\mathbf{T}}_{s,y}$.The similarity calculation aims to quantify the similarity between the data characteristics of this sample and the overall distribution characteristics of each site. The higher the similarity, the more matched the data distribution between this sample and the corresponding site. Each site provides two templates $\bar{\mathbf{T}}_{s,0}$ and $\bar{\mathbf{T}}_{s,1}$, then the original attention score of that site can be written as follows:
\begin{equation}\alpha_{s} = \cos(T,\bar{\mathbf{T}}_{s,0}) + \cos(T,\bar{\mathbf{T}}_{s,1}),\end{equation}
Normalizing the original attention scores of each site can obtain the final attention score. The closer the data characteristics are to the target sample, the greater the attention weight obtained by the site, and the higher its proportion in the final decision. The weight of predicted logits for each site is as follows
\begin{equation}w_{l,s}=\frac{\alpha_{s}}{\sum_{j=1}^{N}\alpha_{j}},\end{equation}

Multiply the attention scores $w_{s}$ of each site by their corresponding predicted logits $\mathbf{l}_s$ and add them together to obtain the final predicted logits $\hat{\mathbf{l}}$, and the calculation process can be formalized as follows:
\begin{equation}\hat{\mathbf{l}}=\sum_{s=1}^{N}w_{l,s}\mathbf{l}_s,\end{equation}
Based on the characteristics of the target sample, dynamically fuse model information from all sites, prioritize the selection and fusion of model decisions from sites with the most similar data distribution to the target sample, effectively alleviate the distribution offset caused by data heterogeneity, and significantly improve the accuracy and robustness of model.

Compare the difference between predicted labels and real labels to calculate the accuracy of the test. This process is independently executed for each input sample, achieving highly personalized prediction.

In this stage, the matching degree between each test sample of the site and the data characteristics of each site is calculated, and attention scores are generated. The final prediction logits is generated by weighted aggregation based on the attention scores, achieving personalized prediction at the single sample level and effectively improving the prediction accuracy.
\begin{table}[hb!]
  \small
  \centering
  \caption{Structures of 4 heterogeneous CNN models}
  \label{tab:cnn-structures}
  \setlength{\tabcolsep}{3pt}   
  \begin{tabular}{lcccc}
    \toprule
     \textbf{Layer} & \textbf{CNN-1} & \textbf{CNN-2} & \textbf{CNN-3} & \textbf{CNN-4}\\
     \midrule
     Conv1 & 1024,\,$1\times$116 & 512,\,$1\times$116 & 1000,\,$1\times$116 & 1024,\,1$\times$116\\
     Conv2 & 2000,\,116$\times$1 & 2000,\,116$\times$1 & 2000,\,116$\times$1 & 2048,\,116$\times$1\\
     Linear1 & 96 & 96  & 96 & 96\\
     Dropout1 & 0.5 & 0.5 & 0.5 & 0.5\\
     Linear2 & 2 & 2 & 2 & 2\\
    \bottomrule
  \end{tabular}
\end{table}
\section{Results}
\subsection{Experiment Setup}
\textbf{Datasets.} We use the REST-meta-MDD dataset\cite{restmdd}, a large-scale multi-center resting-state fMRI collection from 17 Chinese hospitals, comprising 1,300 patients with MDD and 1,128 healthy controls across 25 cohorts. This multi-site design captures real-world data heterogeneity, ideal for evaluating federated learning under diverse clinical conditions.

All sites applied the same preprocessing pipeline using DPARSF\footnote{\url{https://rfmri.org/REST-meta-MDD}}: slice timing, motion correction, normalization (MNI152), nuisance regression, and temporal filtering (0.01–0.1 Hz). To mitigate motion artifacts, the Friston-24-parameter model was regressed out per subject, and mean frame displacement was included as a covariate in group analyses.

To investigate neurobiological differences across MDD subtypes, we stratified patients from the REST-meta-MDD dataset using two clinically relevant dimensions: \textit{episodicity} (first vs. recurrent episode) and \textit{medication status} (medicated vs. drug-naive). Combining these factors, we defined four subgroups with distinct clinical profiles:
\begin{itemize}
\item \textbf{Site 1}: Recurrent MDD, unmedicated (n=76)
\item \textbf{Site 2}: Recurrent MDD, medicated (n=121)
\item \textbf{Site 3}: First-episode, drug-naïve (FEDN) (n=318)
\item \textbf{Site 4}: First-episode, medicated (n=160)
\end{itemize}
To ensure balanced classification tasks, we matched each patient group with an equal number of healthy controls (HCs) randomly sampled from the NC cohort. The resulting sites have the following total sizes: Site 1: 152 (76 MDD + 76 NC), Site 2: 242 (121 + 121), Site 3: 636 (318 + 318), Site 4: 320 (160 + 160).

\textbf{Base Models.}
Two models are set for the classifier to evaluate the impact of model heterogeneity in experiments. The first type is heterogeneous classifiers for each site, which is closer to the scenario where different models are adopted due to differences in computing power, data distribution, or business requirements in reality. Four heterogeneous CNNs are assigned to each of the four sites. The second type is that the classifiers of each site are homogeneous, and all sites uniformly use CNN-1 in Table 2 as the local classifier.
To evaluate the impact of model heterogeneity, we design two experimental settings for the local classifiers:

\textbf{(1) Heterogeneous classifiers}: Reflecting real-world variations in computational resources, data distribution, and clinical requirements, each site uses a distinct CNN architecture. Four different CNNs are assigned to the four sites, allowing personalized modeling capacity.

\textbf{(2) Homogeneous classifiers}: All sites adopt the same architecture, specifically \texttt{CNN-1} (see Table 2), enabling a controlled comparison to assess the effect of architectural diversity.

This setup allows us to analyze how model heterogeneity influences performance and robustness in federated brain image classification under clinical subgroup diversity.

\subsection{Comparisons With Other Methods}
To evaluate the effectiveness of our AAA framework, we compare it with three established federated learning methods: FedAvg~\cite{2016}, FedProto~\cite{fedproto}, and pFedAFM~\cite{pfedafm}. Although these baselines were originally evaluated on benchmarks like CIFAR-10 and MNIST, we re-implement and re-train them on the REST-meta-MDD dataset using the same data partitioning and preprocessing pipeline to ensure a fair comparison.

All methods are tested under both homogeneous and heterogeneous classifier settings , and performance is evaluated site-wise. The classification accuracy for each method and site is reported in Table 3.

\begin{table*}[htb]
  \centering
  \caption{Accuracy of methods in the field of federated learning}
  \label{tab:FL-acc}
  \begin{tabular}{lcccccc}
    \toprule
    Field & Method        & Site1 & Site2 & Site3 & Site4 & Average\\
    \midrule
    \multirow{5}{*}{\makecell{Federated \\ Learning}} & FedAvg     & 0.7419 & 0.6327 & 0.6875 & 0.7188 & 0.6952 \\
            & FedProto   & 0.6452 & 0.7347 & 0.6172 & 0.6016 & 0.6497 \\
            & pFedAFM    & 0.7750 & 0.6111 & 0.6404 & 0.7214 & 0.6870 \\
            & Ours (homo)   & 0.8500 & 0.6311 & 0.6769 & 0.7571 & 0.7288 \\
            & \textbf{Ours (hetero)} & 0.8500 & 0.6933 & 0.6904 & 0.7286 & \textbf{0.7406} \\
    \bottomrule
  \end{tabular}
\end{table*}

The following provides a detailed introduction to these three comparison methods:
\begin{table*}[ht!]
  \centering
  \caption{ACC of the Ablation Study in the MDD Diagnostic Classification}
  \label{tab:Ablation}
  \begin{tabular}{lcccccc}
    \toprule
    Subset & MoE        & Site1 & Site2 & Site3 & Site4 & Average\\
    \midrule
    \multirow{2}{*}{$\checkmark$ } & $\checkmark$     & 0.8500 & 0.6933 & 0.6904 & 0.7286 & \textbf{0.7406} \\
            & $\times$   & 0.8000 & 0.6733 & 0.6538 & 0.7214 & 0.7121 \\
    \multirow{2}{*}{$\times$ }        & $\checkmark$    & 0.8250 & 0.6311 & 0.6269 & 0.6786 & 0.6904 \\
            & $\times$  & 0.7750 & 0.5933 & 0.6077 & 0.6357 & 0.6529 \\
    \bottomrule
  \end{tabular}
\end{table*}
\begin{enumerate}
    \item The FedAvg model\cite{fedavg}, which represents the FDL model trained with the federated gradient averaging learning method.
    \item The Fedproto model\cite{fedproto}, which represents federated prototype learning across heterogeneous clients.
    \item The Pfedafm model\cite{pfedafm}, which represents heterogeneous federated learning for implementing batch personalized adaptive feature mixing.
\end{enumerate}

In the field of deep learning, we also investigated three methods, all of which were conducted on the REST-meta-MDD, and we directly calculated the ACC and AUC of these three methods. The results are shown in Figure 4.

\begin{figure}[htb]
    \centering
    \includegraphics[scale=0.3]{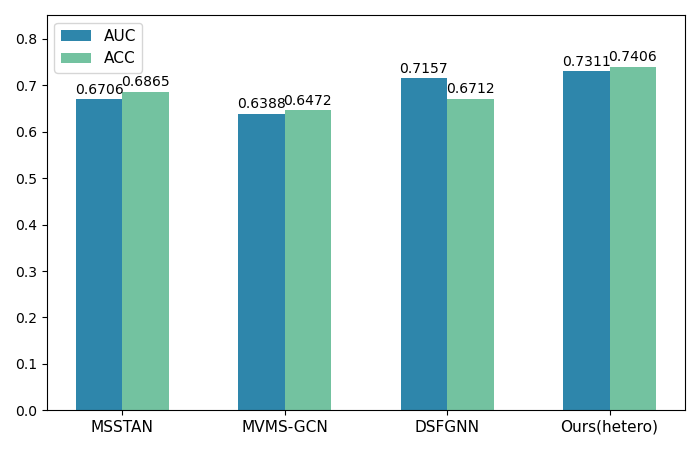} 
    \caption{Acc and Auc of methods in the field of deep learning}
    \label{fig:relative-path}
\end{figure}

The following provides a detailed introduction to these three comparison methods:

\begin{enumerate}
    \item The MSSTAN model\cite{MSSTAN}, which represents a multi view multi-source data fusion graph convolutional neural network.
    \item The MVMS-GCN model\cite{MVMS-GCN}, which represents a multi graph spectral brain network based on spectral neural network, is used for fusion diagnosis of severe depression.
    \item The DSFGNN model\cite{DSFGNN}, which represents a dynamic static fusion graph neural network, is used to enhance the diagnosis of depression.
\end{enumerate}





As shown in Table 2 and Figure 4, AAA achieves the highest average classification accuracy among all evaluated methods. Compared to the best-performing federated learning baseline, AAA improves accuracy by 4.54\%; against the top deep learning baseline, the improvement reaches 5.41\%. These consistent gains demonstrate the effectiveness of our framework in leveraging subtype-specific patterns for personalized prediction. The results indicate that incorporating clinical subtyping and adaptive aggregation enables AAA to better handle data and model heterogeneity, leading to significantly enhanced performance in multi-site medical image classification.

\subsection{Ablation Study}
To evaluate the contribution of key components in the proposed AAA framework, we conduct an ablation study by comparing four variants in MDD diagnostic classification. The complete model incorporates both subtype-based grouping and the MoE mechanism for dynamic, personalized classifier fusion. We then consider a variant that retains subtype classification but removes the MoE module, relying instead on hard selection of the most similar site. A third variant disables subtype grouping (treating all patients uniformly) but keeps the MoE for adaptive fusion. Finally, we evaluate a baseline that removes both subtype classification and MoE, resulting in a static aggregation without personalization. This setup allows us to isolate the effects of clinical subtyping and adaptive attention aggregation on model performance.


As shown in Table 4, the complete AAA model—incorporating both subtype classification and the MoE-based dynamic fusion—achieves the highest performance among all ablation variants. It outperforms the model with subtype classification but without MoE, as well as the model that uses MoE without subtype stratification. This consistent improvement demonstrates that both components contribute independently and positively to classification accuracy. The results suggest that clinical subtype modeling provides a meaningful patient stratification, while the MoE module enables more adaptive and personalized prediction by dynamically weighting local models. Together, they enable a more effective federated framework for heterogeneous medical data.
\section{Conclusion}
Our study demonstrates that the proposed AAA framework, which integrates subtype classification with a MoE module for dynamic, personalized classifier fusion, achieves superior performance in multi-site MDD diagnostic classification. By leveraging clinical subtypes and adaptive aggregation, the complete model significantly outperforms baseline methods that either omit subtype stratification or lack the MoE mechanism. Specifically, our approach improves accuracy by 4.54\% over the best federated learning baseline and by 5.41\% over the best deep learning baseline. These findings highlight the importance of incorporating both patient-specific characteristics and adaptive modeling strategies to enhance predictive performance in heterogeneous datasets. Future work will focus on extending this framework to other psychiatric disorders and exploring additional methods for improving robustness and generalizability across diverse clinical settings.

\bigskip

\bibliography{AnonymousSubmission/fed}

\begin{thebibliography}{32}
\providecommand{\natexlab}[1]{#1}

\bibitem[{Fang et~al.(2023)Fang, Wang, Potter, and Liu}]{domdd}
Fang, Y.; Wang, M.; Potter, G.~G.; and Liu, M. 2023.
\newblock Unsupervised cross-domain functional MRI adaptation for automated major depressive disorder identification.
\newblock \emph{Medical Image Analysis}, 84: 102707.

\bibitem[{Guan et~al.(2024)Guan, Yap, Bozoki, and Liu}]{fedsurvey}
Guan, H.; Yap, P.-T.; Bozoki, A.; and Liu, M. 2024.
\newblock Federated learning for medical image analysis: A survey.
\newblock \emph{PATTERN RECOGNITION}, 151.

\bibitem[{Hu et~al.(2023)Hu, Huang, Liu, Xue, Sun, Song, and Tan}]{dotpami}
Hu, Y.; Huang, Z.-A.; Liu, R.; Xue, X.; Sun, X.; Song, L.; and Tan, K.~C. 2023.
\newblock Source Free Semi-Supervised Transfer Learning for Diagnosis of Mental Disorders on fMRI Scans.
\newblock \emph{IEEE Transactions on Pattern Analysis and Machine Intelligence}, 45(11): 13778--13795.

\bibitem[{Jiang et~al.(2024)Jiang, Le, Li, and Dou}]{pfedes}
Jiang, M.; Le, A.; Li, X.; and Dou, Q. 2024.
\newblock Heterogeneous Personalized Federated Learning by Local-Global Updates Mixing via Convergence Rate.
\newblock In \emph{The Twelfth International Conference on Learning Representations}.

\bibitem[{Kairouz et~al.(2021)Kairouz, McMahan, Avent, Bellet, Bennis, Nitin~Bhagoji, Bonawitz, Charles, Cormode, Cummings, D’Oliveira, Eichner, El~Rouayheb, Evans, Gardner, Garrett, Gasc\'{o}n, Ghazi, Gibbons, Gruteser, Harchaoui, He, He, Huo, Hutchinson, Hsu, Jaggi, Javidi, Joshi, Khodak, Konecn\'{y}, Korolova, Koushanfar, Koyejo, Lepoint, Liu, Mittal, Mohri, Nock, \"{O}zg\"{u}r, Pagh, Qi, Ramage, Raskar, Raykova, Song, Song, Stich, Sun, Suresh, Tram\`{e}r, Vepakomma, Wang, Xiong, Xu, Yang, Yu, Yu, and Zhao}]{FL}
Kairouz, P.; McMahan, H.~B.; Avent, B.; Bellet, A.; Bennis, M.; Nitin~Bhagoji, A.; Bonawitz, K.; Charles, Z.; Cormode, G.; Cummings, R.; D’Oliveira, R. G.~L.; Eichner, H.; El~Rouayheb, S.; Evans, D.; Gardner, J.; Garrett, Z.; Gasc\'{o}n, A.; Ghazi, B.; Gibbons, P.~B.; Gruteser, M.; Harchaoui, Z.; He, C.; He, L.; Huo, Z.; Hutchinson, B.; Hsu, J.; Jaggi, M.; Javidi, T.; Joshi, G.; Khodak, M.; Konecn\'{y}, J.; Korolova, A.; Koushanfar, F.; Koyejo, S.; Lepoint, T.; Liu, Y.; Mittal, P.; Mohri, M.; Nock, R.; \"{O}zg\"{u}r, A.; Pagh, R.; Qi, H.; Ramage, D.; Raskar, R.; Raykova, M.; Song, D.; Song, W.; Stich, S.~U.; Sun, Z.; Suresh, A.~T.; Tram\`{e}r, F.; Vepakomma, P.; Wang, J.; Xiong, L.; Xu, Z.; Yang, Q.; Yu, F.~X.; Yu, H.; and Zhao, S. 2021.
\newblock Advances and Open Problems in Federated Learning.
\newblock \emph{Found. Trends Mach. Learn.}, 14(1–2): 1–210.

\bibitem[{Kone{\v c}n{\'y} et~al.(2016)Kone{\v c}n{\'y}, McMahan, Yu, Richt{\'a}rik, Suresh, and Bacon}]{2016}
Kone{\v c}n{\'y}, J.; McMahan, H.; Yu, F.; Richt{\'a}rik, P.; Suresh, A.; and Bacon, D. 2016.
\newblock Federated Learning: Strategies for Improving Communication Efficiency: Strategies for Improving Communication Efficiency.
\newblock Workingpaper, ArXiv.

\bibitem[{Kong et~al.(2025)Kong, Zhang, Wang, Zhou, Li, and Yuan}]{MSSTAN}
Kong, Y.; Zhang, X.; Wang, W.; Zhou, Y.; Li, Y.; and Yuan, Y. 2025.
\newblock Multi-Scale Spatial-Temporal Attention Networks for Functional Connectome Classification.
\newblock \emph{IEEE TRANSACTIONS ON MEDICAL IMAGING}, 44(1): 475--488.

\bibitem[{Li et~al.(2020)Li, Gu, Dvornek, Staib, Ventola, and Duncan}]{muti}
Li, X.; Gu, Y.; Dvornek, N.; Staib, L.~H.; Ventola, P.; and Duncan, J.~S. 2020.
\newblock Multi-site fMRI analysis using privacy-preserving federated learning and domain adaptation: ABIDE results.
\newblock \emph{Medical image analysis}, 65: Article 101765.

\bibitem[{Marek et~al.(2022)Marek, Tervo-Clemmens, Calabro, Montez, Kay, Hatoum, Donohue, Foran, Miller, Hendrickson, Malone, Kandala, Feczko, Miranda-Dominguez, Graham, Earl, Perrone, Cordova, Doyle, Moore, Conan, Uriarte, Snider, Lynch, Wilgenbusch, Pengo, Tam, Chen, Newbold, Zheng, Seider, Van, Metoki, Chauvin, Laumann, Greene, Petersen, Garavan, Thompson, Nichols, Yeo, Barch, Luna, Fair, and Dosenbach}]{moresamples}
Marek, S.; Tervo-Clemmens, B.; Calabro, F.~J.; Montez, D.~F.; Kay, B.~P.; Hatoum, A.~S.; Donohue, M.~R.; Foran, W.; Miller, R.~L.; Hendrickson, T.~J.; Malone, S.~M.; Kandala, S.; Feczko, E.; Miranda-Dominguez, O.; Graham, A.~M.; Earl, E.~A.; Perrone, A.~J.; Cordova, M.; Doyle, O.; Moore, L.~A.; Conan, G.~M.; Uriarte, J.; Snider, K.; Lynch, B.~J.; Wilgenbusch, J.~C.; Pengo, T.; Tam, A.; Chen, J.; Newbold, D.~J.; Zheng, A.; Seider, N.~A.; Van, A.~N.; Metoki, A.; Chauvin, R.~J.; Laumann, T.~O.; Greene, D.~J.; Petersen, S.~E.; Garavan, H.; Thompson, W.~K.; Nichols, T.~E.; Yeo, B. T.~T.; Barch, D.~M.; Luna, B.; Fair, D.~A.; and Dosenbach, N. U.~F. 2022.
\newblock Reproducible brain-wide association studies require thousands of individuals.
\newblock \emph{NATURE}, 603(7902): 654+.

\bibitem[{McMahan et~al.(2016)McMahan, Moore, Ramage, Hampson, and y~Arcas}]{fedavg}
McMahan, H.~B.; Moore, E.; Ramage, D.; Hampson, S.; and y~Arcas, B.~A. 2016.
\newblock Communication-Efficient Learning of Deep Networks from Decentralized Data.
\newblock In \emph{International Conference on Artificial Intelligence and Statistics}.

\bibitem[{Qin et~al.(2022)Qin, Lei, Pinaya, Pan, Li, Zhu, Sweeney, Mechelli, and Gong}]{mdd}
Qin, K.; Lei, D.; Pinaya, W. H.~L.; Pan, N.; Li, W.; Zhu, Z.; Sweeney, J.~A.; Mechelli, A.; and Gong, Q. 2022.
\newblock Using graph convolutional network to characterize individuals with major depressive disorder across multiple imaging sites.
\newblock \emph{EBIOMEDICINE}, 78.

\bibitem[{Radua et~al.(2020)Radua, Vieta, Shinohara, Kochunov, Quide, Green, Weickert, Weickert, Bruggemann, Kircher, Nenadic, Cairns, Seal, Schall, Henskens, Fullerton, Mowry, Pantelis, Lenroot, Cropley, Loughland, Scott, Wolf, Satterthwaite, Tan, Sim, Piras, Spalletta, Banaj, Pomarol-Clotet, Solanes, Albajes-Eizagirre, Canales-Rodriguez, Sarro, Di~Giorgio, Bertolino, Staeblein, Oertel, Knoechel, Borgwardt, du~Plessis, Yun, Kwon, Dannlowski, Hahn, Grotegerd, Alloza, Arango, Janssen, Diaz-Caneja, Jiang, Calhoun, Ehrlich, Yang, Cascella, Takayanagi, Sawa, Tomyshev, Lebedeva, Kaleda, Kirschner, Hoschl, Tomecek, Skoch, van Amelsvoort, Bakker, James, Preda, Weideman, Stein, Howells, Uhlmann, Temmingh, Lopez-Jaramillo, Diaz-Zuluaga, Fortea, Martinez-Heras, Solana, Llufriu, Jahanshad, Thompson, Turner, van Erp, and Collaborators}]{hetero}
Radua, J.; Vieta, E.; Shinohara, R.; Kochunov, P.; Quide, Y.; Green, M.~J.; Weickert, C.~S.; Weickert, T.; Bruggemann, J.; Kircher, T.; Nenadic, I.; Cairns, M.~J.; Seal, M.; Schall, U.; Henskens, F.; Fullerton, J.~M.; Mowry, B.; Pantelis, C.; Lenroot, R.; Cropley, V.; Loughland, C.; Scott, R.; Wolf, D.; Satterthwaite, T.~D.; Tan, Y.; Sim, K.; Piras, F.; Spalletta, G.; Banaj, N.; Pomarol-Clotet, E.; Solanes, A.; Albajes-Eizagirre, A.; Canales-Rodriguez, E.~J.; Sarro, S.; Di~Giorgio, A.; Bertolino, A.; Staeblein, M.; Oertel, V.; Knoechel, C.; Borgwardt, S.; du~Plessis, S.; Yun, J.-Y.; Kwon, J.~S.; Dannlowski, U.; Hahn, T.; Grotegerd, D.; Alloza, C.; Arango, C.; Janssen, J.; Diaz-Caneja, C.; Jiang, W.; Calhoun, V.; Ehrlich, S.; Yang, K.; Cascella, N.~G.; Takayanagi, Y.; Sawa, A.; Tomyshev, A.; Lebedeva, I.; Kaleda, V.; Kirschner, M.; Hoschl, C.; Tomecek, D.; Skoch, A.; van Amelsvoort, T.; Bakker, G.; James, A.; Preda, A.; Weideman, A.; Stein, D.~J.; Howells, F.; Uhlmann, A.; Temmingh, H.; Lopez-Jaramillo, C.;
  Diaz-Zuluaga, A.; Fortea, L.; Martinez-Heras, E.; Solana, E.; Llufriu, S.; Jahanshad, N.; Thompson, P.; Turner, J.; van Erp, T.; and Collaborators, E.~C. 2020.
\newblock Increased power by harmonizing structural MRI site differences with the ComBat batch method in ENIGMA.
\newblock \emph{NEUROIMAGE}, 218.

\bibitem[{Richards et~al.(2019)Richards, Lillicrap, Beaudoin, Bengio, Bogacz, Christensen, Clopath, Costa, de~Berker, Ganguli, Gillon, Hafner, Kepecs, Kriegeskorte, Latham, Lindsay, Miller, Naud, Pack, Poirazi, Roelfsema, Sacramento, Saxe, Scellier, Schapiro, Senn, Wayne, Yamins, Zenke, Zylberberg, Therien, and Kording}]{dlinns}
Richards, B.~A.; Lillicrap, T.~P.; Beaudoin, P.; Bengio, Y.; Bogacz, R.; Christensen, A.; Clopath, C.; Costa, R.~P.; de~Berker, A.; Ganguli, S.; Gillon, C.~J.; Hafner, D.; Kepecs, A.; Kriegeskorte, N.; Latham, P.; Lindsay, G.~W.; Miller, K.~D.; Naud, R.; Pack, C.~C.; Poirazi, P.; Roelfsema, P.; Sacramento, J.; Saxe, A.; Scellier, B.; Schapiro, A.~C.; Senn, W.; Wayne, G.; Yamins, D.; Zenke, F.; Zylberberg, J.; Therien, D.; and Kording, K.~P. 2019.
\newblock A deep learning framework for neuroscience.
\newblock \emph{NATURE NEUROSCIENCE}, 22(11): 1761--1770.

\bibitem[{Rieke et~al.(2020)Rieke, Hancox, Li, Milletari, Roth, Albarqouni, Bakas, Galtier, Landman, Maier-Hein, Ourselin, Sheller, Summers, Trask, Xu, Baust, and Cardoso}]{fedhealth}
Rieke, N.; Hancox, J.; Li, W.; Milletari, F.; Roth, H.~R.; Albarqouni, S.; Bakas, S.; Galtier, M.~N.; Landman, B.~A.; Maier-Hein, K.; Ourselin, S.; Sheller, M.; Summers, R.~M.; Trask, A.; Xu, D.; Baust, M.; and Cardoso, M.~J. 2020.
\newblock The future of digital health with federated learning.
\newblock \emph{NPJ DIGITAL MEDICINE}, 3(1).

\bibitem[{Shi et~al.(2023{\natexlab{a}})Shi, Yi, Liu, and Wang}]{ffedcl}
Shi, X.; Yi, L.; Liu, X.; and Wang, G. 2023{\natexlab{a}}.
\newblock FFEDCL: Fair Federated Learning with Contrastive Learning.
\newblock In \emph{ICASSP 2023 - 2023 IEEE International Conference on Acoustics, Speech and Signal Processing (ICASSP)}, 1--5.

\bibitem[{Shi et~al.(2023{\natexlab{b}})Shi, Yao, Yi, Yu, Zhang, and Li}]{fedwm}
Shi, Z.; Yao, Z.; Yi, L.; Yu, H.; Zhang, L.; and Li, X.-Y. 2023{\natexlab{b}}.
\newblock FedWM: Federated Crowdsourcing Workforce Management Service for Productive Laziness.
\newblock In \emph{2023 IEEE International Conference on Web Services (ICWS)}, 152--160.

\bibitem[{Shinn et~al.(2023)Shinn, Hu, Turner, Noble, Preller, Ji, Moujaes, Achard, Scheinost, Constable, Krystal, Vollenweider, Lee, Anticevic, Bullmore, and Murray}]{fcST}
Shinn, M.; Hu, A.; Turner, L.; Noble, S.; Preller, K.~H.; Ji, J.~L.; Moujaes, F.; Achard, S.; Scheinost, D.; Constable, R.~T.; Krystal, J.~H.; Vollenweider, F.~X.; Lee, D.; Anticevic, A.; Bullmore, E.~T.; and Murray, J.~D. 2023.
\newblock Functional brain networks reflect spatial and temporal autocorrelation.
\newblock \emph{NATURE NEUROSCIENCE}, 26(5).

\bibitem[{Smith et~al.(2021)Smith, Douaud, Chen, Hanayik, Alfaro-Almagro, Sharp, and Elliott}]{datasets}
Smith, S.~M.; Douaud, G.; Chen, W.; Hanayik, T.; Alfaro-Almagro, F.; Sharp, K.; and Elliott, L.~T. 2021.
\newblock An expanded set of genome-wide association studies of brain imaging phenotypes in UK Biobank.
\newblock \emph{NATURE NEUROSCIENCE}, 24(5): 737--745.

\bibitem[{Smith and Nichols(2018)}]{bigdata}
Smith, S.~M.; and Nichols, T.~E. 2018.
\newblock Statistical Challenges in ``Big Data'' Human Neuroimaging.
\newblock \emph{NEURON}, 97(2): 263--268.

\bibitem[{Song et~al.(2023)Song, Zhou, Frangi, Cao, Xiao, Lei, Wang, and Lei}]{nonimage}
Song, X.; Zhou, F.; Frangi, A.~F.; Cao, J.; Xiao, X.; Lei, Y.; Wang, T.; and Lei, B. 2023.
\newblock Multicenter and Multichannel Pooling GCN for Early AD Diagnosis Based on Dual-Modality Fused Brain Network.
\newblock \emph{IEEE Transactions on Medical Imaging}, 42(2): 354--367.

\bibitem[{Tan et~al.(2022)Tan, Long, Liu, Zhou, Lu, Jiang, and Zhang}]{fedproto}
Tan, Y.; Long, G.; Liu, L.; Zhou, T.; Lu, Q.; Jiang, J.; and Zhang, C. 2022.
\newblock FedProto: Federated Prototype Learning across Heterogeneous Clients.
\newblock AAAI Conference on Artificial Intelligence, 8432--8440.

\bibitem[{Wachinger et~al.(2021)Wachinger, Rieckmann, Poelsterl, Ini, and Li}]{multisite}
Wachinger, C.; Rieckmann, A.; Poelsterl, S.; Ini, A. D.~N.; and Li, A. I. B.~. 2021.
\newblock Detect and correct bias in multi-site neuroimaging datasets.
\newblock \emph{MEDICAL IMAGE ANALYSIS}, 67.

\bibitem[{Wang, Chen, and Yan(2023)}]{covbat}
Wang, Y.-W.; Chen, X.; and Yan, C.-G. 2023.
\newblock Comprehensive evaluation of harmonization on functional brain imaging for multisite data-fusion.
\newblock \emph{NeuroImage}, 274: 120089.

\bibitem[{Yan et~al.(2022)Yan, Chen, Li, Castellanos, Bai, Bo, Cao, Chen, Chen, Chen, Cheng, Cheng, Cui, Duan, Fang, Gong, Guo, Hou, Hu, Kuang, Li, Li, Liu, Liu, Long, Luo, Meng, Peng, Qiu, Qiu, Shen, Shi, Tang, Wang, Wang, Wang, Wang, Wang, Wang, Wu, Wu, Xie, Xie, Xie, Xie, Xu, Yang, Yang, Yao, Yao, Yin, Yuan, Zhang, Zhang, Zhang, Zhang, Zhang, Zhou, Zhou, Zhu, Zou, Si, Zuo, Zhao, and Zang}]{restmdd}
Yan, C.-G.; Chen, X.; Li, L.; Castellanos, F.~X.; Bai, T.-J.; Bo, Q.-J.; Cao, J.; Chen, G.-M.; Chen, N.-X.; Chen, W.; Cheng, C.; Cheng, Y.-Q.; Cui, X.-L.; Duan, J.; Fang, Y.-R.; Gong, Q.-Y.; Guo, W.-B.; Hou, Z.-H.; Hu, L.; Kuang, L.; Li, F.; Li, T.; Liu, Y.-S.; Liu, Z.-N.; Long, Y.-C.; Luo, Q.-H.; Meng, H.-Q.; Peng, D.-H.; Qiu, H.-T.; Qiu, J.; Shen, Y.-D.; Shi, Y.-S.; Tang, Y.-Q.; Wang, C.-Y.; Wang, F.; Wang, K.; Wang, L.; Wang, X.; Wang, Y.; Wu, X.-P.; Wu, X.-R.; Xie, C.-M.; Xie, G.-R.; Xie, H.-Y.; Xie, P.; Xu, X.-F.; Yang, H.; Yang, J.; Yao, J.-S.; Yao, S.-Q.; Yin, Y.-Y.; Yuan, Y.-G.; Zhang, A.-X.; Zhang, H.; Zhang, K.-R.; Zhang, L.; Zhang, Z.-J.; Zhou, R.-B.; Zhou, Y.-T.; Zhu, J.-J.; Zou, C.-J.; Si, T.-M.; Zuo, X.-N.; Zhao, J.-P.; and Zang, Y.-F. 2022.
\newblock {Data of the REST-meta-MDD Project from DIRECT Consortium}.

\bibitem[{Yang et~al.(2019)Yang, Liu, Chen, and Tong}]{mode}
Yang, Q.; Liu, Y.; Chen, T.; and Tong, Y. 2019.
\newblock Federated Machine Learning: Concept and Applications.
\newblock 10(2).

\bibitem[{Yi et~al.(2023)Yi, Wang, Liu, Shi, and Yu}]{fedgh}
Yi, L.; Wang, G.; Liu, X.; Shi, Z.; and Yu, H. 2023.
\newblock FedGH: Heterogeneous Federated Learning with Generalized Global Header.
\newblock In \emph{Proceedings of the 31st ACM International Conference on Multimedia}, MM '23, 8686–8696. New York, NY, USA: Association for Computing Machinery.
\newblock ISBN 9798400701085.

\bibitem[{Yi et~al.(2024)Yi, Yu, Shi, Wang, Liu, Cui, and Li}]{fedssa}
Yi, L.; Yu, H.; Shi, Z.; Wang, G.; Liu, X.; Cui, L.; and Li, X. 2024.
\newblock FedSSA: semantic similarity-based aggregation for efficient model-heterogeneous personalized federated learning.
\newblock In \emph{Proceedings of the Thirty-Third International Joint Conference on Artificial Intelligence}, IJCAI '24.
\newblock ISBN 978-1-956792-04-1.

\bibitem[{Yi et~al.(2025)Yi, Yu, Wang, Liu, and Li}]{pfedafm}
Yi, L.; Yu, H.; Wang, G.; Liu, X.; and Li, X. 2025.
\newblock { pFedAFM: Adaptive Feature Mixture for Data-Level Personalization in Heterogeneous Federated Learning on Mobile Edge Devices }.
\newblock In \emph{2025 IEEE 41st International Conference on Data Engineering (ICDE)}, 1981--1994. Los Alamitos, CA, USA: IEEE Computer Society.

\bibitem[{Zeng et~al.(2024)Zeng, Fan, Su, Gan, Peng, Shen, and Hu}]{fedbrainimage}
Zeng, L.-L.; Fan, Z.; Su, J.; Gan, M.; Peng, L.; Shen, H.; and Hu, D. 2024.
\newblock Gradient Matching Federated Domain Adaptation for Brain Image Classification.
\newblock \emph{IEEE TRANSACTIONS ON NEURAL NETWORKS AND LEARNING SYSTEMS}, 35(6): 7405--7419.

\bibitem[{Zhai et~al.(2024)Zhai, Liu, Xiong, Qu, and Yang}]{MVMS-GCN}
Zhai, K.; Liu, Q.; Xiong, Y.; Qu, X.; and Yang, J. 2024.
\newblock MVMS-GCN: A Multi-view Multi-source data fusion graph convolution neural network for predicting autism spectrum disorder with fMRI.
\newblock In \emph{PROCEEDINGS OF 2024 4TH INTERNATIONAL CONFERENCE ON BIOINFORMATICS AND INTELLIGENT COMPUTING, BIC 2024}, 26--31.
\newblock ISBN 979-8-4007-1664-5.
\newblock 4th International Conference on Bioinformatics and Intelligent Computing (BIC), Beijing, PEOPLES R CHINA, JAN 26-28, 2024.

\bibitem[{Zhao, Adeli, and Pohl(2020)}]{confounderfree}
Zhao, Q.; Adeli, E.; and Pohl, K.~M. 2020.
\newblock Training confounder-free deep learning models for medical applications.
\newblock \emph{NATURE COMMUNICATIONS}, 11(1).

\bibitem[{Zhao and Zhang(2024)}]{DSFGNN}
Zhao, T.; and Zhang, G. 2024.
\newblock Enhancing Major Depressive Disorder Diagnosis With Dynamic-Static Fusion Graph Neural Networks.
\newblock \emph{IEEE JOURNAL OF BIOMEDICAL AND HEALTH INFORMATICS}, 28(8): 4701--4710.

\end{thebibliography}
\newpage
\clearpage
\end{document}